\documentclass[letterpaper, 10pt, conference]{IEEEtran}
\usepackage{amsmath}
\usepackage{amssymb}
\usepackage{color}
\usepackage{graphicx}
\usepackage{url}
\usepackage{algorithm}
\usepackage{algorithmic}
\usepackage{tikz}
\usepackage{mathtools}
\usepackage{colortbl}
\usepackage{tabularx,booktabs}
\usepackage{caption, subcaption, stfloats}
\usepackage[backend=biber, style=ieee, sorting=none]{biblatex}
\begin{document}

\title{Enhancing Graph-Based SLAM in GNSS-Denied environments by leveraging leg odometry}
\author{
    \IEEEauthorblockN{Léon PERRUCHOT-TRIBOULET\IEEEauthorrefmark{1}, 
                      Luc JAULIN\IEEEauthorrefmark{1} and 
                      Kai XIAO\IEEEauthorrefmark{2}}
    \IEEEauthorblockA{\IEEEauthorrefmark{1}Lab-STICC, ENSTA Campus Brest, France\\
    Email: \{leon.perruchot-triboulet, luc.jaulin\}@ensta.fr}
    \IEEEauthorblockA{\IEEEauthorrefmark{2}LinXai Tech. Co., China\\
    Email: xiaokai@linxai-tech.com}
}
\maketitle

\begin{abstract}
Autonomous navigation in GNSS-denied environments remains a core challenge for legged robots, where exteroceptive sensors such as LiDAR are prone to elevation drift in geometrically sparse or repetitive scenes.
We present a factor graph architecture that augments the LIO-SAM framework with a parallel kinematic lane driven by proprioceptive leg odometry, coupled to the main LiDAR-inertial lane via an identity relative pose constraint with a selective noise model.
Applied to a Linxai D50 quadruped platform across two outdoor loops totaling over one kilometer, our approach reduces elevation drift from over 30m to under 30cm and enables convergence in a scene where the baseline pipeline fails entirely.
These results suggest that proprioceptive data, already computed onboard for gait control, constitutes a lightweight and effective vertical anchor for SLAM in GNSS-denied settings.
\end{abstract}

\begin{figure*}[!tb]
    \centering
    \begin{subfigure}[b]{0.45\textwidth}
        \centering
        \includegraphics[width=\textwidth]{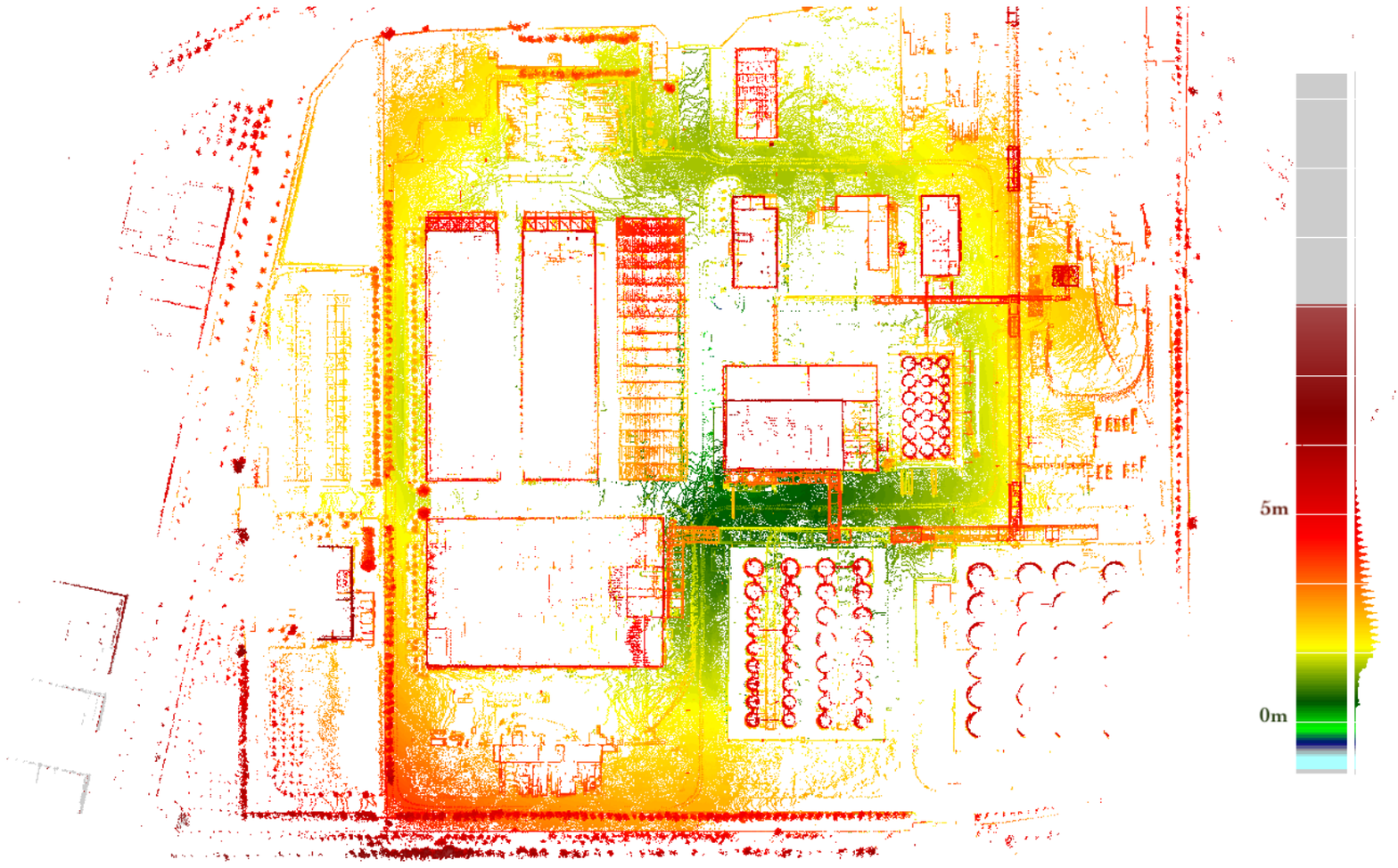}
        \caption{Factory Dataset}
        \label{fig:map_factory}
    \end{subfigure}
    \hfill
    \begin{subfigure}[b]{0.45\textwidth}
        \centering
        \includegraphics[width=\textwidth]{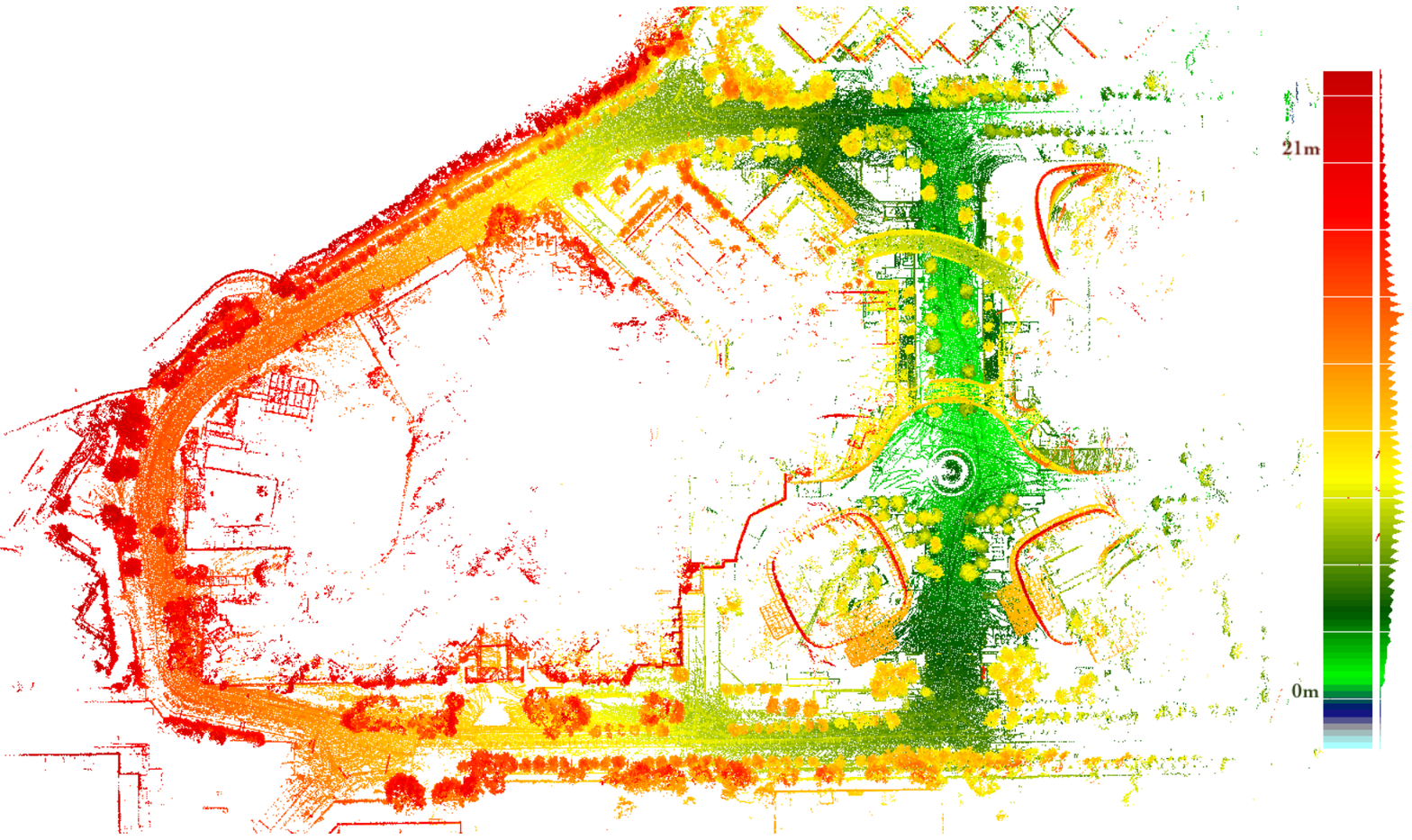}
        \caption{CocoPark Dataset}
        \label{fig:map_cocopark}
    \end{subfigure}
    \caption{Map generated with the proposed modified SLAM framework on both experimental datasets. Each image is independently colormapped according to the elevation, with green being lower and red higher.}
    \label{fig:all_maps}
\end{figure*}

\section{Introduction}
Simultaneous Localization and Mapping (SLAM) is a prerequisite for long duration autonomous operation of mobile robots. In GNSS-denied environments (tunnels, dense urban streets, or GPS-degraded outdoor scenes) LiDAR-inertial systems are among the most capable available, combining scan-to-map matching with IMU preintegration in a factor graph that supports loop closure~\cite{liosam}.
However, these systems are vulnerable to a specific and practically significant failure mode: elevation drift. When a scene lacks distinctive vertical geometry (open plazas, long roads, foliage-bordered paths) the ICP frontend provides weak constraints along the z-axis, and small per frame biases accumulate into large global errors.
LIO-SAM natively addresses this through the addition of GNSS factors. In the absence of GNSS, this anchor is simply absent.
Legged robots offer a natural remedy. Their locomotion inherently encodes vertical displacement, each foot ground contact constrains the body height relative to the terrain, and forward kinematics (FK) provides a continuous 6-DoF relative motion estimate from joint encoder data already collected for gait control. This proprioceptive signal is geometrically independent of the exteroceptive scene, making it complementary to LiDAR precisely where LiDAR is weak.
This paper makes the following contributions: in Section \ref{sec:relatedWork} we identify the integration architecture that allows leg odometry to act as a vertical regularizer within LIO-SAM's factor graph without degrading horizontal accuracy. Section \ref{sec:approach} introduces a parallel kinematic lane coupled to the main LiDAR lane via an identity relative pose constraint with a selective noise model. Finally, Section \ref{sec:experiments} demonstrates on two outdoor missions a reduction in elevation error from over 30m to under 30cm, including successful convergence in a scene where the baseline fails entirely.

\section{Related Work}
\label{sec:relatedWork}
\subsection{Graph-based LiDAR-inertial SLAM}
Modern SLAM systems increasingly rely on factor graphs for multi-sensor fusion~\cite{lego-loam, slam}. iSAM2~\cite{isam2} introduced incremental Bayes-tree smoothing, enabling near-constant-time updates and exposed via the GTSAM library~\cite{gtsam}. LIO-SAM~\cite{liosam} builds on this formalism by tightly coupling LiDAR scan matching with IMU preintegration, with support for loop closure and GNSS factors.
Our work uses LIO-SAM as a base framework and extends its factor graph without modifying any internal component. We support this choice for two main reasons. Firstly LIO-SAM being a well-studied SLAM framework, small modifications made to the algorithm can easily be compared with other's work. Secondly this choice is made for ease of development as the D50 robot (see \ref{subsec:system}) used in our tests comes with LIO-SAM.

\subsection{Leg odometry for state estimation}
Proprioceptive state estimation on legged robots has a well-established literature. Bloesch et al.~\cite{FK_stateEstimation} fused FK and IMU via an EKF, and provided an initial leap into the observability of a legged robot's state. VILENS~\cite{VILENS} demonstrated tight coupling of leg odometry, IMU, camera and LiDAR in a factor graph for outdoor autonomy on ANYmal. Compared with these works, we focus specifically on the elevation axis and on minimally invasive integration into an existing production SLAM pipeline, without modifying the robot's low-level controller.

Our work is positioned as a lightweight, deployment oriented contribution: no custom sensor, no changes to the existing software stack, and a design that degrades gracefully. If the kinematic lane is disabled, the system falls back to standard LIO-SAM.

\section{Approach}
\label{sec:approach}
\subsection{System Overview}
\label{subsec:system}
The experimental platform is the D50 quadruped (LinxAI Tech), equipped with a 16 ring 3D LiDAR, a 500 Hz IMU, and twelve joint encoders (3 per limb).
All modules communicate over ROS Noetic. The FK odometry pipeline was already present on the platform as part of the low-level gait controller, continuously publishing a 6-DoF odometry estimate that we reuse without modification.
LIO-SAM operates at two frequencies: a 500 Hz IMU preintegration layer, and a 20 Hz map optimization layer where scan matching, and optional GNSS factors are inserted. Our contribution operates entirely at the 20 Hz optimization layer.

\subsection{Failure Mode of the baseline}
\begin{figure}[!b]
    \centering
    \includegraphics[width=0.9\linewidth]{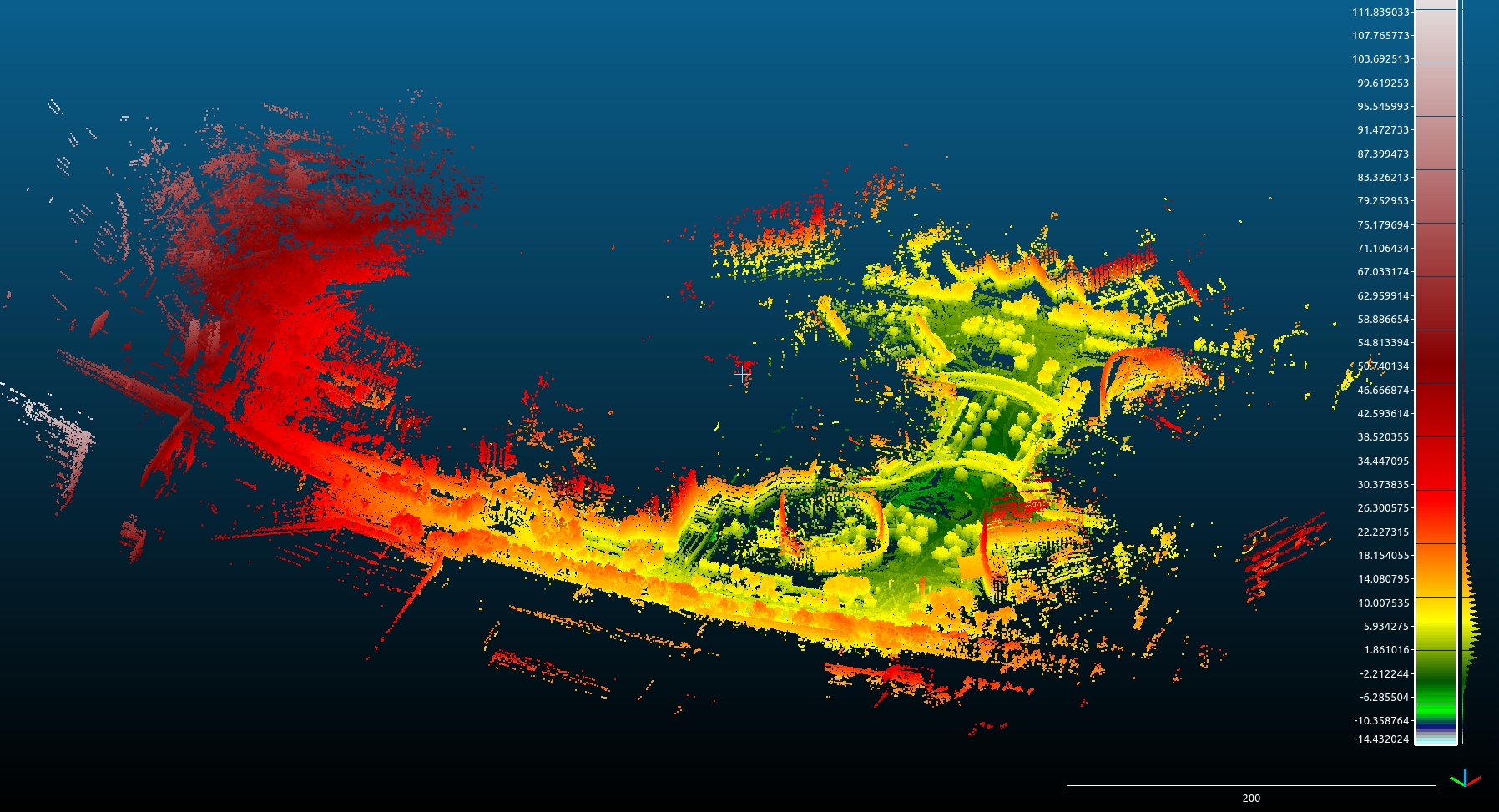}
    \caption{SLAM pipeline crash with visible false elevation etimates. Map generated running the baseline LIO-SAM algorithm on the D50 robot (without GNSS) in the CocoPark environment.}
    \label{fig:map_crash}
\end{figure}

Without a GNSS factor, the baseline LIO-SAM on the D50 exhibits two failure modes.
First, elevation drift: in scenes with sparse vertical structure, small z biases accumulate, producing a characteristic map-wrapping artifact with over 30 meters of apparent elevation gain on a loop with less than 5 meters of true relief along a 400 meters long path.
Second, divergence: in scenes with dynamic objects and adversarial geometry (foliage, bike parkings), accumulated pose uncertainty causes the ICP search radius to be exceeded and the pipeline crashes with no partial map recovered. Figure \ref{fig:map_crash} illustrates a pipeline crash.
Both failure modes motivate a sensor that can provide a vertical anchor unconditionally and independently of scene geometry.

\subsection{Hybrid Factor Graph Architecture}
We evaluated two integration strategies:
Serial integration, inserting the FK odometry factor directly into the existing LIO-SAM graph alongside the LiDAR and IMU factors, and parallel kinematic lane, a separate subgraph coupled via a soft constraint. \\
Serial integration proved ineffective: the LiDAR–IMU factors already tightly constrain all six DoFs, and an additional factor on the same node pair introduces conflicting horizontal information without improving elevation. All tested serial variants produced negligible elevation improvement.
The parallel architecture avoids this by keeping each sensor's constraints in a dedicated subgraph where it is authoritative.

\paragraph{LiDAR-Inertial lane}
The original LIO-SAM factor graph is retained without modification (except for the absence of GNSS factors and loop-closure constraints). It provides accurate horizontal localization through scan to map ICP matching and IMU preintegration.

\paragraph{Forward kinematics lane}
A second, independent subgraph runs in parallel.
Its nodes $\mathbf{y}_k \in SE(3)$ represent pose estimates derived exclusively from leg odometry. It contains prior factors anchoring the elevation of each keyframe relative to the local ground surface extracted from FK, exploiting the fact that foot-ground contact directly constrains body height above terrain; and relative pose factors encoding the 6-DoF relative displacement between consecutive keyframes as estimated by the on board FK pipeline. Leg odometry values are linearly interpolated to align with LiDAR keyframe timestamps.

\begin{figure}
    \centering
    \includegraphics[width=\linewidth]{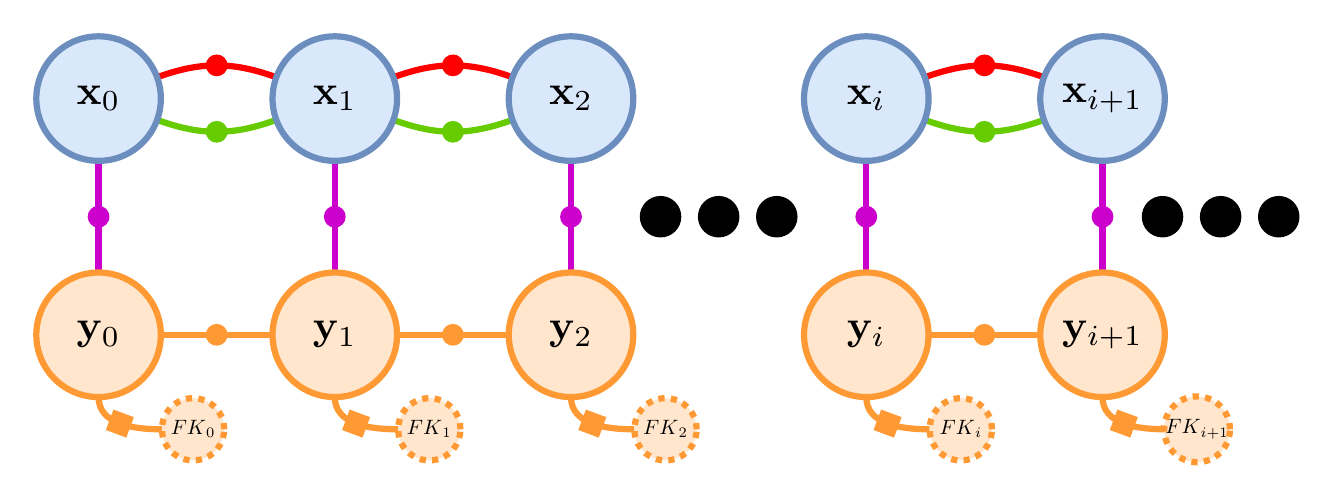}
    \caption{Hybrid factor graph with a parallel lane leg-odometry state estimation. In blue circles LiDAR-inertial state estimates, in orange circles leg-odometry state estimates, the rounded red lines Imu preintegration factor, the rounded green lines LiDAR scan-to-map odometry relative pose factors, the rounded orange lines relative leg-odometry relative pose factors, the squared orange lines leg-odometry elevation prior factors and the purple lines soft identity contraint.} 
    \label{fig:factor_graph}
\end{figure}

\paragraph{Cross-lane coupling via relative pose factors}
The two lanes share corresponding pose nodes $\mathbf{x}_k$ (LiDAR lane) and $\mathbf{y}_k$ (kinematic lane) at each keyframe.
They are coupled by an identity factor with GTSAM's \texttt{BetweenFactor<Pose3>} factor and a cost function expressed by equation \ref{eq:cost}.
\begin{equation}
\label{eq:cost}
    f_{coupling}(\mathbf{x}_k, \mathbf{y}_k) = \|log(\mathbf{x}_k^{-1} \mathbf{y}_k)\|^{2}_{\Sigma}
\end{equation}
With $\|\cdot\|^2_{\Sigma}$ the squared Mahalanobis norm, and $log$ the mapping function from the Lie group manifold to the Lie algebra.
The noise model $\Sigma$ is a diagonal matrix with tight variance on z (trusting the kinematic lane on elevation) and large variance on the other state variables (trusting the LiDAR lane on horizontal pose and orientation). This encodes the complementary reliability of each sensor directly in the factor graph structure. The combined graph is optimized with GTSAM using the iSAM2 incremental solver.
A diagram of the proposed factor graph structure is illustrated in Figure \ref{fig:factor_graph}

\paragraph{Output and asymmetry}
The final published pose estimate is sampled from the LiDAR lane nodes $\mathbf{x}_k$ only. The kinematic lane acts as a soft vertical regularizer, it pulls the LiDAR-inertial lane nodes toward elevation consistent solutions during joint optimization, but the horizontal accuracy properties of LIO-SAM are fully preserved. This is an important design distinction, the system is not a symmetric sensor fusion but an asymmetric regularization of the LiDAR estimate via the coupled kinematic subgraph.

\section{Experiments}
\label{sec:experiments}
\subsection{Setup}
Experiments were conducted over two teleoperated outdoor missions on the D50 platform, with all data recorded as ROS bags then processed offline.
Ground truth via GNSS was unavailable on this platform at the time, we report loop closure discrepancy i.e., the start to finish pose error at the end of each closed loop as the primary metric.
Both missions form single large loops with limited intermediate loop closure opportunities. Closures occur only near the end of each trajectory when the robot returns close to its starting point.
Table \ref{tab:datasets} summarizes the key aspects of each dataset.`

\begin{table}
    \centering
    \caption{Experimental dataset characteristics}
    \label{tab:datasets}
    \footnotesize
    \begin{tabularx}{\columnwidth}{lXX}
        \toprule
        \textbf{Dataset} & Factory & CocoPark \\
        \midrule
        \textbf{Length} & $\sim$700\,m & $\sim$600\,m \\
        \addlinespace
        \textbf{Relief} & $<5$\,m (flat) & $\sim$15\,m (hilly) \\
        \addlinespace
        \textbf{Key challenges} & repetitive facades & dynamic objects, foliage, bike racks \\
        \bottomrule
    \end{tabularx}
\end{table}

\subsection{Quantitative Results}
Results are summarized in Table \ref{tab:results}, all errors are loop-closure discrepancies, path lengths are approximate and "diverged" indicates the baseline crashed before completing the route.
On the Factory dataset, the baseline produced a complete but severely distorted map, with over 30 meters of elevation error and 5 meters of horizontal error at loop closure.
The proposed hybrid graph reduced elevation error to 20 cm and horizontal error to approximately 2 meters, confirming that the kinematic lane does not degrade horizontal accuracy.
On the CocoPark dataset, the baseline diverged and crashed before completing the loop. With leg-odometry regularization, the pipeline converged over the full route, achieving 30 cm elevation error and approximately 4 meters horizontal error at loop closure.

\begin{table}[htb]
    \centering
    \caption{Loop-closure discrepancy (m) for both datasets}
    \label{tab:results}
    \footnotesize
    \begin{tabularx}{\columnwidth}{llXX}
        \toprule
        \textbf{Dataset} & \textbf{Method} & \textbf{$\Delta z$} & \textbf{$\Delta xy$} \\
        \midrule
        Factory  & Baseline & $>30$\,m & $\sim$5\,m \\
                 & \textbf{Proposed} & \textbf{0.2\,m} & \textbf{$\sim$2\,m} \\
        \addlinespace
        CocoPark & Baseline & diverged & diverged \\
                 & \textbf{Proposed} & \textbf{0.3\,m} & \textbf{$\sim$4\,m} \\
        \bottomrule
    \end{tabularx}
\end{table}

\subsection{Qualitative Observations}
The hybrid approach eliminates the map-wrapping artifact in all tested scenes, producing globally consistent 3D point clouds that visually align with satellite imagery of both sites. A secondary benefit was observed: anchoring the elevation estimate reduced pose uncertainty propagated to the ICP front-end, improving scan matching stability in the horizontal plane as well.

\section{Discussions}
\label{sec:discussions}
In this paper we applied vertical regularization, not full fusion. The published output is from the LiDAR-inertial lane only. The leg odometry lane constrains the factor graph via the relative pose factor coupling, but it does not directly contribute to the horizontal pose estimates. This preserves LIO-SAM's horizontal accuracy by design, but also means that FK odometry provides limited benefit on the horizontal axis in scenarios where LiDAR is degraded, although it appears it helped the convergence of the factor graph optimization. \par
A symmetric fusion, where FK contributes to all degrees of freedom with contact quality weighted covariance, is a natural extension while avoiding the over constraining effect observed in our serial integration experiments. \par
Our evaluation uses loop closure discrepancy as a proxy for accuracy. On these datasets (single large loops with closures only at the end), the metric captures integrated drift over the full mission but does not characterize mid trajectory accuracy. An independent ground truth source (e.g. RTK-GNSS) would be beneficial to ensure good estimates and is a priority for future work. \par
The kinematic lane assumes FK odometry is locally reliable between consecutive keyframes. The current implementation has no explicit slip detection. Under severe slip, prior factors in the kinematic lane could introduce incorrect anchors.
The current approach is a practical proof of concept, a future direction is to build a dedicated FK factor from raw joint angles to remove the dependency on the gait controller's internal odometry.
However, in practice this did not appear to degrade results on these datasets (robot walking solely on dry asphalt ground), but a covariance model conditioned on contact quality remains a promising avenue. \par
All results are from one robot and one SLAM framework. Generalization to other legged platforms or other LiDAR-inertial systems (FAST-LIO2~\cite{fastLIO2}, LiLi-OM~\cite{liliom}) is architecturally straightforward, the parallel lane design being framework agnostic, but has not been experimentally validated.

\section{Conclusion and Future Work}
\label{sec:conclusion}
We have presented a hybrid factor graph architecture that integrates proprioceptive leg odometry as a parallel kinematic lane within LIO-SAM, coupled to the LiDAR-inertial lane via an identity relative pose constraint with a DoF-selective noise model. Applied to a D50 quadruped across two outdoor closed-loop missions, the approach reduces elevation drift from over 30 meters to under 30 centimeters and enables convergence where the baseline pipeline fails without modifying the existing SLAM pipeline or the robot's low level controller.
The key insight is architectural: serial insertion of FK odometry into an existing tightly coupled factor graph over constrains the problem and yields negligible benefit, whereas a parallel lane coupled by a soft elevation selective factor preserves each sensor's authority in its reliable degrees of freedom.
Future directions include: symmetric fusion with contact quality weighted covariance on all axes; integration of dynamic object segmentation (e.g., MotionSeg3D) before the ICP front end and evaluation against an independent ground.

\printbibliography

\end{document}